# MedMobile: A mobile-sized language model with clinical capabilities

Krithik Vishwanath[1-3], Jaden Stryker[1], Anton Alyakin[1,4], Daniel Alexander Alber[1], Eric Karl Oermann[1,5,6]

[1]Department of Neurological Surgery, NYU Langone Medical Center, New York, New York, USA
[2]Department of Aerospace Engineering & Engineering Mechanics, The University of Texas at Austin, Austin, Texas, USA
[3]Department of Mathematics, The University of Texas at Austin, Austin, Texas, USA
[4]Department of Neurosurgery, Washington University School of Medicine in St. Louis, St. Louis, Missouri, USA
[5]Department of Radiology, NYU Langone Health, New York, New York, USA
[6]Center for Data Science, New York University, New York, New York, USA

**Correspondence:**
Krithik Vishwanath
Department of Neurosurgery,
NYU Langone Medical Center,
New York University, 550 First Ave, MS 3 205,
New York, NY10016, USA.
Email: krithik.vish@utexas.edu

Eric K. Oermann, MD
Department of Neurosurgery,
NYU Langone Medical Center,
New York University, 550 First Ave, MS 3 205,
New York, NY10016, USA.
Email: eric.oermann@nyulangone.org

**GitHub:** https://github.com/nyuolab/MedMobile

**Preprint: Accepted @ BMJ Digital Health & AI**

**STRUCTURED ABSTRACT**

**OBJECTIVE:** Language models (LMs) have demonstrated expert-level reasoning and recall abilities in medicine. However, computational costs and privacy concerns are mounting barriers to wide-scale implementation. To address these significant limitations, we introduce a parsimonious adaptation of phi-3-mini, MedMobile, a 3.8 billion parameter LM capable of running on a mobile device, for medical applications.

**METHODS AND ANALYSIS:** We perform a careful set of pipeline additions and demonstrate that chain of thought, ensembling, and fine-tuning lead to the greatest performance gains, while unexpectedly retrieval augmented generation fails to demonstrate significant improvements. We evaluate the efficiency of our pipeline on the MultiMedQA and Medbullets.

**RESULTS:** We demonstrate that MedMobile scores 75.7% on the MedQA (USMLE), surpassing the passing mark for licensed physicians (~60%) and rivaling scores of models 100 times its size. Across the entirety of the MultiMedQA, MedMobile achieves SOTA performance for models with less than 5B parameters and represents the smallest model to pass the MedQA (USMLE).

**CONCLUSIONS:** MedMobile holds promise to democratize access to language models in medicine, bolstering lower compute needs and fast inference speeds. With the ability to combat the biggest barriers to entry for language models in medicine, we hope that MedMobile is a critical step forward in developing clinically relevant language models.

## Introduction

In recent years, language models (LMs) have revolutionized numerous tasks in natural language processing, from question answering and summarization to dialogue and decision support. In the medical domain specifically, LMs have shown notable promise for accelerating clinical workflows, whether by rapidly retrieving relevant literature, generating draft reports, or assisting in differential diagnosis through reasoning over complex patient presentations [1–3]. Their ability to encode vast amounts of biomedical knowledge and apply it flexibly makes them attractive tools for clinicians and healthcare systems striving to improve both efficiency and quality of care [4].

Despite these advances, large-scale adoption of LMs in real-world medical settings remains limited by several practical and technical barriers. First, serving and maintaining state-of-the-art models entails substantial computational costs: inference on multi-hundred-billion-parameter architectures demands high-performance GPUs or specialized hardware, which many hospitals and clinics cannot afford [5,6]. Second, concerns around data privacy and security are acute in healthcare, where patient information must be rigorously protected; deploying models that communicate with external APIs can run afoul of regulatory requirements. Third, the most capable LMs today (such as GPT-3.5 and GPT-4) are closed-source, preventing independent auditing, domain-specific fine-tuning, and integration into local, on-premises systems [7]. These factors combine to create a landscape in which the promise of LMs in medicine is often out of reach for those who might benefit most. These challenges are exponentiated in low-resource and remote settings.

To navigate these challenges, we explore a middle ground by leveraging phi-3-mini, an open-source language model with 3.8 billion parameters. At roughly two orders of magnitude smaller than GPT-3.5 (~175 billion parameters), phi-3-mini strikes a balance between expressive capacity and deployability on resource-constrained hardware [8]. We fine-tune this base model on a specialized medical corpus composed of (i) high-quality, expert-curated question-answer pairs, and (ii) synthetic examples generated by prompting GPT-4 with medical textbooks and guidelines. This hybrid approach harnesses the precision of human annotation alongside the breadth and creativity of large-model generation, enabling the smaller model to acquire domain-specific reasoning skills without incurring prohibitive data-collection costs.

The resulting model, which we call MedMobile, is designed for lightweight, on-device inference. This opens the door to point-of-care applications on tablets, smartphones, or embedded clinical workstations. In evaluations on MedQA, a benchmark of USMLE-style medical exam questions, MedMobile achieves an overall accuracy of 75.7%, substantially surpassing the ~60% "passing" threshold and marking it, to our knowledge, as the smallest open-source LM to meet or exceed this standard [9]. Through MedMobile, we demonstrate that carefully fine-tuned, small-sized

LMs can deliver clinically relevant performance at a fraction of the computational cost and with the transparency required for medical deployment.

Enabling smaller language models to achieve superior performance on the USMLE-style and other medical tasks is an active area of research [10,11]. Due to advancements in language model architecture, higher quality training data, and novel prompt engineering techniques, recent open-source models in the ~7-8B range, such as Meerkat [10] and UltraMedical Llama 3.1 [11], have achieved a passing score on USMLE-style Q&A, even outperforming language models several times larger such as GPT-3.5 (175B) [12], the SOTA from two years ago. Meerkat [10], the first 7B parameter model to achieve such a distinction, focused on improving smaller models via synthetic textbook-based, USMLE-style questions generated by GPT-4. Another series of models, UltraMedical [11] expands this work, generating synthetic questions on a larger scale and across all question types in the MultiMedQA [13]. Generalist language models can be improved significantly with knowledge distillation via supervised fine-tuning (SFT) on synthetic data. In this regard, enhancing smaller language models with support from much larger models have emerged as a leading approach to achieving superior performance with low-compute requirements.

Since there is a significant loss of token generation speed and increase in power consumption on mobile devices for models sizes larger than 5B parameters [14], we use the terminology "mobile-size" to refer to LMs smaller than 5B parameters. In this context, MedMobile is the first mobile-sized model to pass the USMLE. To achieve this with a reduced computational burden, we choose phi-3-mini as a backbone for our model as it exhibits enhanced reasoning capabilities relative to other models of the size [8]. Models that exhibit superior reasoning often utilize Chain-of-Thought (CoT), a technique that emulates human-like reasoning by using a chain of logical steps to achieve an accurate conclusion. State-of-the-art language models, such as GPT-4, expose a clear step-by-step process for arriving at its answer, thereby shedding insight into the reasoning process that models may employ. By fine-tuning phi-3-mini using this CoT of GPT-4 (i.e., the logical process it uses to achieve its final conclusion), we retain generalist reasoning capabilities while gaining medical domain knowledge, partially distilling the advanced problem-solving process and knowledge of GPT-4 to phi-3-mini [15]. MedMobile marks a clinically significant advancement by enabling individuals to carry a board-certified clinical assistant in their pocket.

## Methods

### Evaluation

To determine an LM's ability in the medical domain, we evaluate the model on the MultiMedQA, a multi-dataset of medical questions [13]. In accordance with most literature on clinical language models and for the sake of standardization, we perform all evaluations on multiple-choice questions (MCQs). The MultiMedQA is composed of 8 individual datasets ranging from USMLE-style questions (MedQA) to College Biology (MMLU College Biology) and is outlined in Table 1. We chose to evaluate on these datasets due to the expert level of medical reasoning and knowledge required for USMLE-style questions and to test the model's ability against the range of medical tasks with the other datasets. Testing on the PubMedQA also demonstrates MedMobile's ability to perform on research-related medical inquiries. In addition to the MultiMedQA, we also evaluate on Medbullets [16]. This addition evaluation is purposed to test a better range of clinical tasks, especially since the MedQA is primarily samples from the USMLE Step 1 [9]. Further, the MedQA test set, due to its prominence in testing medical models, has a high likelihood to be contaminated in the training of most current-day language models, so this alternative benchmark may serve as a more agnostic method of evaluation. To calculate accuracy, we use string-based matching for CoT and use logarithmic probability for non-CoT experiments. Inference is computed at a temperature of 0 for cases without ensemble, and 0.7 when ensemble is used to allow for differences between model responses within an ensemble. Phi-3-mini and MedMobile are ran using the PyTorch and Transformers library on A100 GPUs during evaluation. We note that utilizing CoT involves a longer response from the model, thus requiring more computational cost and complexity compared to a non-CoT experiment. Per the usage notes on UltraMedical, we utilize vLLM to develop results on medical benchmarks not publicized in literature for their model [11]. Benchmark results for other models are obtained from literature.

**Table 1.** Multiple-choice question (MCQ) datasets used for evaluation, sourced from [13] and [16].

| **Task** | **# of MCQs** | **Task Overview** |
|---|---|---|
| MedQA (USMLE) | 1273 | MCQs based on the US medical licensing exam |
| Medbullets (4-op) | 308 | MCQs based on the US medical licensing exam (Step 2&3) |
| MedMCQA | 4183 | MCQs based on Indian medical entrance exams (NEET, AIIMS) |
| MMLU (Clinical knowledge) | 265 | Clinical knowledge MCQs |
| MMLU (Medical genetics) | 100 | Medical genetics MCQs |
| MMLU (Anatomy) | 135 | Anatomy MCQs |
| MMLU (Professional medicine) | 272 | Professional medicine MCQs (based on USMLE) |
| MMLU (College biology) | 144 | College biology MCQs |
| MMLU (College medicine) | 173 | College medicine MCQs |
| PubMedQA | 500 | Research question with a PubMed abstract |

Pipeline development

In the past few years, several techniques have demonstrated improvement in LMs' Q&A performance on various benchmarks [15,17]. However, we find a lack of technique validation for our context, given that a technique may only be valid for a specific size model, domain, or cocktail of techniques. To determine the positively contributing components of our pipeline, we add components one by one and evaluate after each addition. The specific ordering of the pipeline is done in accordance with common literature, such as the pipeline in MedPrompt [1]. After component testing, we develop our final training pipeline (Figure 1a) built on SFT, prompt optimization, CoT, and response ensembling. We note that proprietary models (i.e., GPT-4), is only utilized during the training of MedMobile, and not necessary during the inference pipeline.

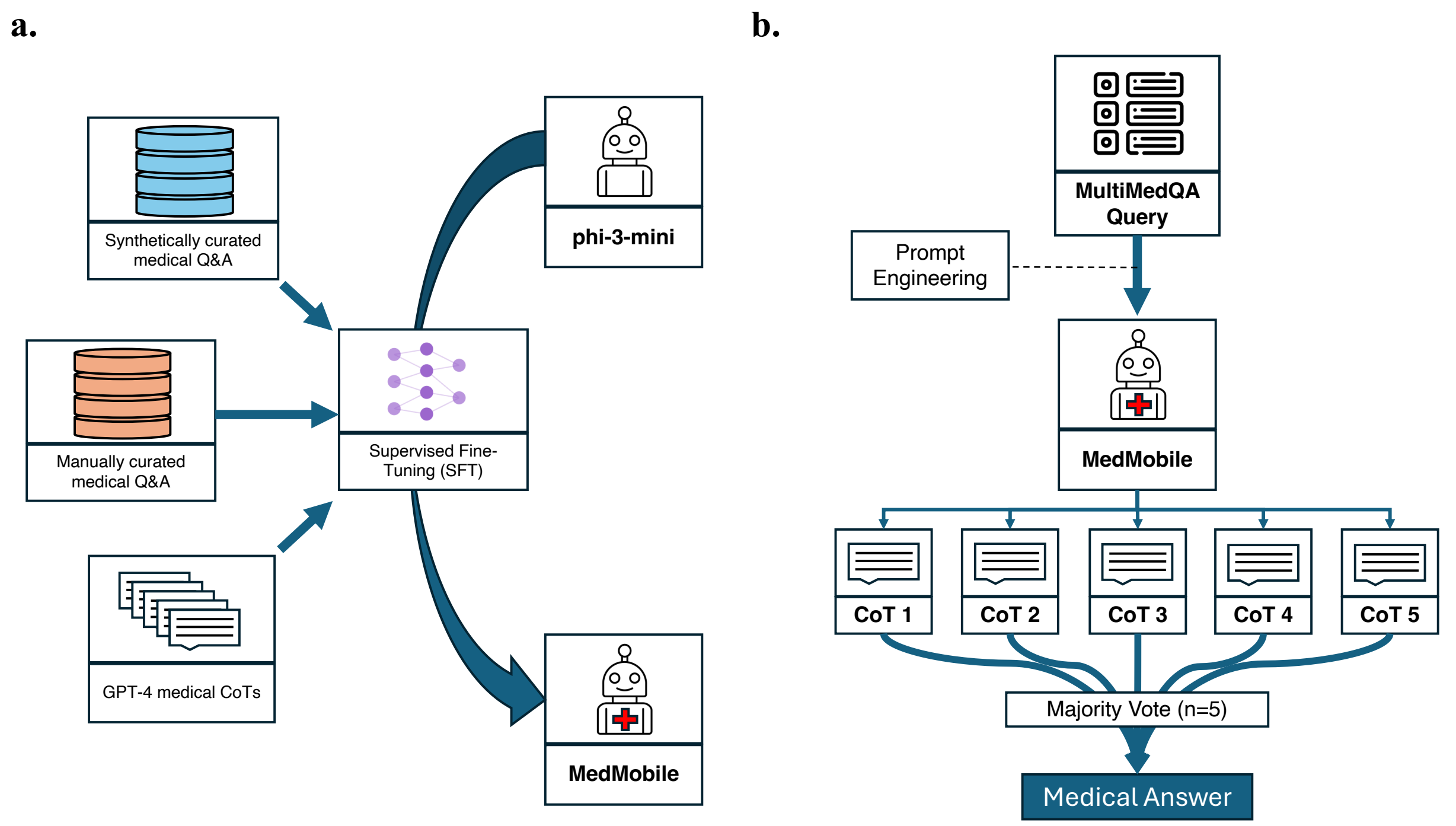

**Figure 1.** Framework of model building and medical Q&A evaluation for MedMobile. **a.** MedMobile is built from the Supervised-Fine Tuning (SFT) of phi-3 mini on the UltraMedical dataset [10], which consists of synthetically curated medical Q&A, manually curated medical Q&A, and GPT-4 Chain-of-Thoughts. **b.** MultiMedQA is used to evaluate MedMobile and uses an optimized prompt engineering template tested using automatic differentiation with GPT-4o as described in TextGrad [17]. Responses are then filtered via an ensemble approach (n=5), where the most common answer is selected as the model's final medical answer.

### Supervised fine-tuning (SFT)

We choose phi-3-mini as our base model due to its demonstrated baseline performance, low parameter count, and training philosophy (i.e., using a combination of high-quality synthetic and human-curated training data) [8]. To train phi-3-mini's baseline parameters to the medical domain, we utilize the UltraMedical dataset, a collection of over 400K synthetic and manual-curated multiple-choice questions [11]. We instruction-fine-tune phi-3-mini using CoT responses from GPT-4 for each of these questions, allowing for knowledge distillation from GPT-4's much larger parameter set. To perform SFT, we train phi-3-mini for 3 epochs on 4 A100 nodes for 83 hours on the UltraMedical dataset. We also utilize a learning rate of $1 \times 10{-}4$ and an effective batch size of 32. A summary of the training details can be found on the GitHub repository and the MedMobile page on HuggingFace.

### Prompt optimization

To ensure streamlined and favorable prompting, we utilize TextGrad [18], a multi-LLM system for improving prompting verbiage using. TextGrad automatically develops improvements to smaller language models' prompts by utilizing a much stronger model (in this case, we use GPT-4). GPT-4o, as an optimizer model, generates new prompting templates. Then, a loss is calculated based on the accuracy generated by MedMobile on the prompt. While TextGrad finds an improved prompt verbiage for phi-3-mini baseline, it also supports that MedMobile does best with no additional prompting instructions. Due to the limited context window capabilities of a model of this size, it is likely that the additional text only hinders the model from domain-specific tasks that it is already trained to reason within. By utilizing the CoT responses of GPT-4 to fine-tune MedMobile, MedMobile exhibits high levels of medical reasoning without the necessity of additional prompting.

Response ensembling

To exploit the stochastic diversity of our small model and boost answer reliability, we adopted a self-consistency ensemble. At inference time we generated 5 independent chain-of-thought (CoT) completions for every question by sampling MedMobile at a temperature of 0.7 (all other hyper-parameters identical to single-shot inference). Each completion was parsed with a rule-based extractor to obtain the model's selected option (e.g., A–E); completions that did not cleanly map to one of the listed choices were recorded as "[invalid]". We tallied the extracted choices across the five attempts and returned the option with the highest frequency as the ensemble answer (majority vote). In the rare case of ties, the first option to reach the maximal count was chosen (Figure 1b).

Other techniques

MedMobile's performance was evaluated with a variety of example-based and retrieval-based prompting strategies to determine whether providing additional context could improve its accuracy on medical multiple-choice questions. In k-shot prompting, the model is supplied with a small set (k) of exemplar question–answer pairs, much like giving a student worked problems before they tackle similar exercises and then asked to answer a new question in the same format. In retrieval-augmented generation (RAG) [19], the model first retrieves potentially relevant passages from an external knowledge source (in our case, selected paragraphs from Harrison's Principles of Internal Medicine, 21e [20]) and uses that information to ground its response.

To compare retrieval methods, we implemented the classic BM25 algorithm (via Lucene) [21], which ranks passages based on the frequency of query terms relative to their rarity (similar to giving extra weight to uncommon "buzzwords" when matching documents) and a cosine-similarity approach in which both questions and candidate paragraphs are embedded with MedCPT [22] and then matched by vector distance. We also tested a hybrid scheme that, for each question, selects whichever passage achieved the highest ranking under either BM25 or embedding similarity.

Despite these varied approaches, the overall accuracy remained statistically unchanged. This suggests that, for this model and dataset, simply adding more or different contextual information does not by itself lead to consistent performance gains.

Estimation of Computational Cost

To approximate the minimum GPU memory required to serve a model, we scale its on-disk parameter size $S_{\text{model}}$ by a constant overhead factor:

$$VRAM_{infer} = 1.2 \times S_{model}.$$

Here, $S_{\text{model}}$ denotes the aggregate size of model weights (in GB) after conversion to a deployment precision (FP16). The multiplicative factor 1.2 heuristically accounts for runtime memory overhead, including (i) framework and kernel workspace allocations, (ii) parameter metadata, and (iii) minimal activation and buffer requirements for batched inference at modest throughput [23]. This simple rule yields a conservative lower bound for single-node, single-process inference. Compute costs for training (ADAM) are approximated as 4× the $VRAM_{infer}$ required for inference [24].

Given a target model, we enumerate feasible hardware configurations and select the least-cost combination that satisfies $\sum_i VRAM_i \geq VRAM_{min}$. The enumeration ("combination algorithm") considers Cartesian products of available GPU instance types and device counts to the minimum needed to meet memory, then filters by compute, ranking surviving configurations by hourly cost.

Pricing is standardized to AWS (us-east-1) on-demand rates (queried at the time of analysis) to ensure transparency and reproducibility. While this procedure captures first-order cost differences across model scales, it slightly underestimates total cost for larger deployments that require tensor, pipeline, or sequence parallelism across multiple instances. In multi-node settings, additional memory (and time) is consumed by (i) activation partitioning and recomputation strategies, (ii) optimizer / gradient state (for training scenarios), and (iii) communication buffers (e.g., NCCL all-reduce workspaces). We do not explicitly model these incremental overheads; thus, reported costs for the largest models should be interpreted as optimistic lower bounds.

Patient and Public Involvement

Patients or the public were not involved in the design, conduct, reporting, or dissemination plans of our research.

## Results

<u>Chain-of-Thought reasoning on USMLE questions</u>

To achieve clinical relevance, we specifically analyze the reasoning outputted by MedMobile. In Figure 2, we present 2 examples of a USMLE question (taken from the MedQA dataset) and the corresponding response from MedMobile. In the first example, the question requires MedMobile to make connections between diagnosis and symptoms. In the second example, we display a MedQA question that requires the tester to identify the best treatment plan given a complex combination of symptoms. In both cases, MedMobile uses chained reasoning similar to GPT-4 to produce the final correct answer. Displayed is only one of five reasoned-out responses to the question, and for evaluation purposes, the most common model response is taken.

### Example Output from *MedMobile*

**Question:** A 55-year-old male visited his primary care physician complaining of chest pain and progressive shortness of breath worsened by exertion. His vital signs were within normal limits. He reports that he worked in a naval shipyard for 10 years during his twenties. A CT scan of his chest shows diffuse thickening of the pleura and a diagnosis of mesothelioma is made. Assuming this diagnosis is correct, which of the following is most likely to also be present in this patient?

A) Pneumothorax B) Pleural effusion C) Systemic inflammatory response syndrome D) Bronchioalveolar carcinoma

The patient's history of working in a naval shipyard and his current symptoms of chest pain and shortness of breath, along with the CT findings of diffuse thickening of the pleura and a diagnosis of mesothelioma, point towards a diagnosis related to asbestos exposure. Mesothelioma . . . is strongly associated with asbestos exposure. One of the common complications of mesothelioma is pleural effusion, which is the accumulation of fluid in the pleural space. This occurs as the disease progresses and can contribute to the patient's symptoms of shortness of breath . . . Given the information provided and the typical complications associated with mesothelioma, pleural effusion is the most likely additional finding in this patient.

Therefore, the answer is [B].

MedMobile Output

**Question:** A 37-year-old woman presents to the emergency department with confusion, anxiety, sweating, and episodes of vomiting. She is accompanied by her sister who says that her symptoms developed acutely and is unsure about what caused it. She is currently on methimazole, which she last took approximately 1 week ago. Approximately 5 days ago she developed an upper respiratory infection managed with bed rest and ibuprofen. Her temperature is 105°F (40.6°C), blood pressure is 95/68 mmHg, pulse is 145/min, and respirations are 23/min. On physical examination, the patient has altered mentation and is agitated and diaphoretic. The patient also has a goiter, exophthalmos, warm skin, and a hand tremor. She is started on intravenous fluids. Which of the following is the best treatment option for this patient?

A) Antibiotics B) Carotid massage C) Clonazepam D) Propranolol

The patient's presentation of confusion, anxiety, sweating, vomiting, fever, tachycardia, low blood pressure, goiter, exophthalmos, warm skin, and hand tremor, in conjunction with her recent history of an upper respiratory infection and the fact that she is on methimazole, is highly suggestive of a thyroid storm. This is a life-threatening exacerbation of hyperthyroidism . . . Propranolol (choice D) is a non-selective beta-blocker that can help control the symptoms of thyroid storm such as tachycardia, anxiety, tremors, and high blood pressure. It also has the added benefit of blocking the peripheral conversion of T4 to the more active T3. Given the clinical scenario and the need for immediate treatment of symptoms associated with thyroid storm, the best treatment option is propranolol.

Therefore, the answer is [D].

MedMobile Output

**Figure 2.** Sample MedQA (USMLE) question and MedMobile's response. Note that this is one of five responses generated before ensembling. MedMobile displays an ability to contextualize complex medical scenarios and develop expert-level conclusions. MedMobile's output is shortened in parts for visual purposes.

Ablation study of pipeline components

In Figure 3, we present an ablation-like study of our pipeline. By adding components one at a time, we hope to discern the more impactful components of our pipeline. We noticed that using self-consistency with 5 responses and using supervised fine-tuning significantly improved model performance. While introducing CoT alone improved model performance, it was not sufficient to surpass the passing threshold on the MedQA. Starting from a baseline phi-3-mini score of 57.5% on the MedQA, adding CoT (+2.4%), ensembling responses (+7.4%), and conducting SFT (+8.4%) allows MedMobile to achieve an accuracy of 75.7%. In conducting our ablation study, we noted several promising potential pipeline components did not favorably impact inference in medical Q&A, such as k-shot prompting with examples (−9.4%) and retrieval-augmented generation (RAG) (−12.6%) from high-quality sources (i.e., textbooks), perhaps due to an increased input token length.

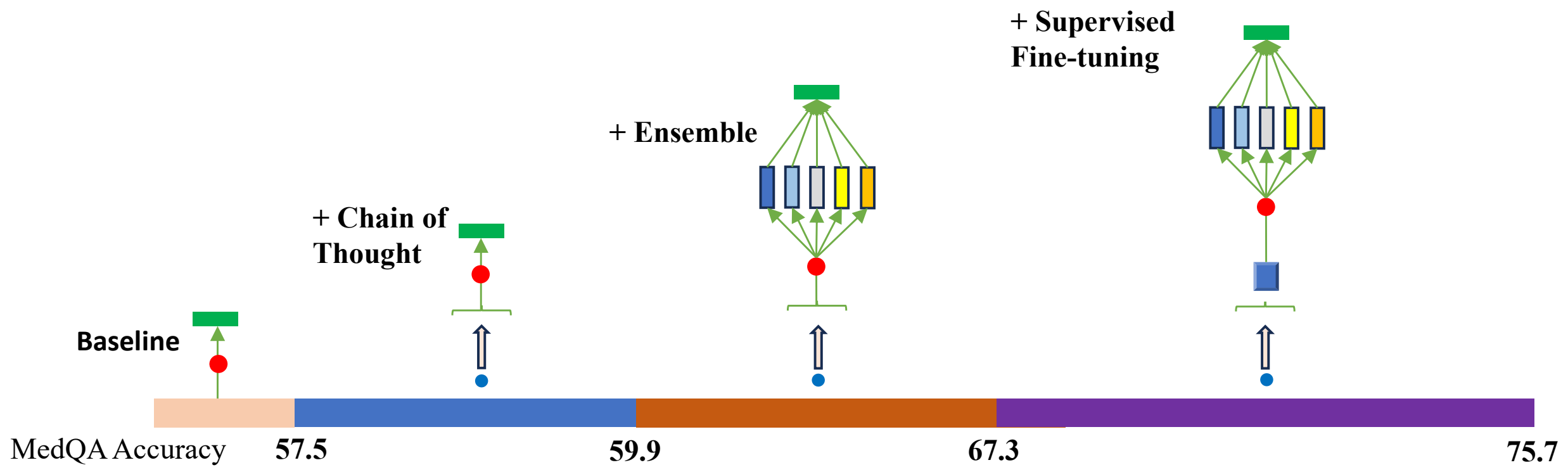

**Figure 3.** Illustration of the MedMobile pipeline and its components by MedQA accuracy. The additions of CoT, self-consistency (ensemble), and supervised fine-tuning in the pipeline lead to significant gains on MedQA accuracy.

State-of-the-Art accuracy with a 3.8B-parameter model

This improvement represents a substantial increase from the next best sub-5B parameter language model, VOD [25], at an accuracy of 55.0% on the MedQA. MedMobile's accuracy on the entirety of the MultiMedQA is comparable to the SOTA models in the medical domain with over double the number of parameters (Figure 4a). In fact, MedMobile beats or matches UltraMedical 8B [10], the current model with the highest accuracy in the sub-10B parameter space, in 6 out of 10 evaluation tasks in the evaluation benchmarks (Figure 4b). To the best of our knowledge, MedMobile is also the smallest model to achieve the distinction of passing USMLE-like questions on the MedQA. Since December 2022, the smallest medical "expert" language model has significantly decreased in parameter size. Flan-PaLM, the first model to achieve this distinction, had over 500B parameters (Figure 4c). MedMobile, at only 3.8B parameters, represents a large reduction in parameters while still maintaining this status of passing the USMLE (based on MedQA performance). Supplemental Table 1 depicts evaluation the raw results for four models. MedMobile and UltraMedical both demonstrate superior performance to Flan-PaLM, despite being a small fraction of the parameter size.

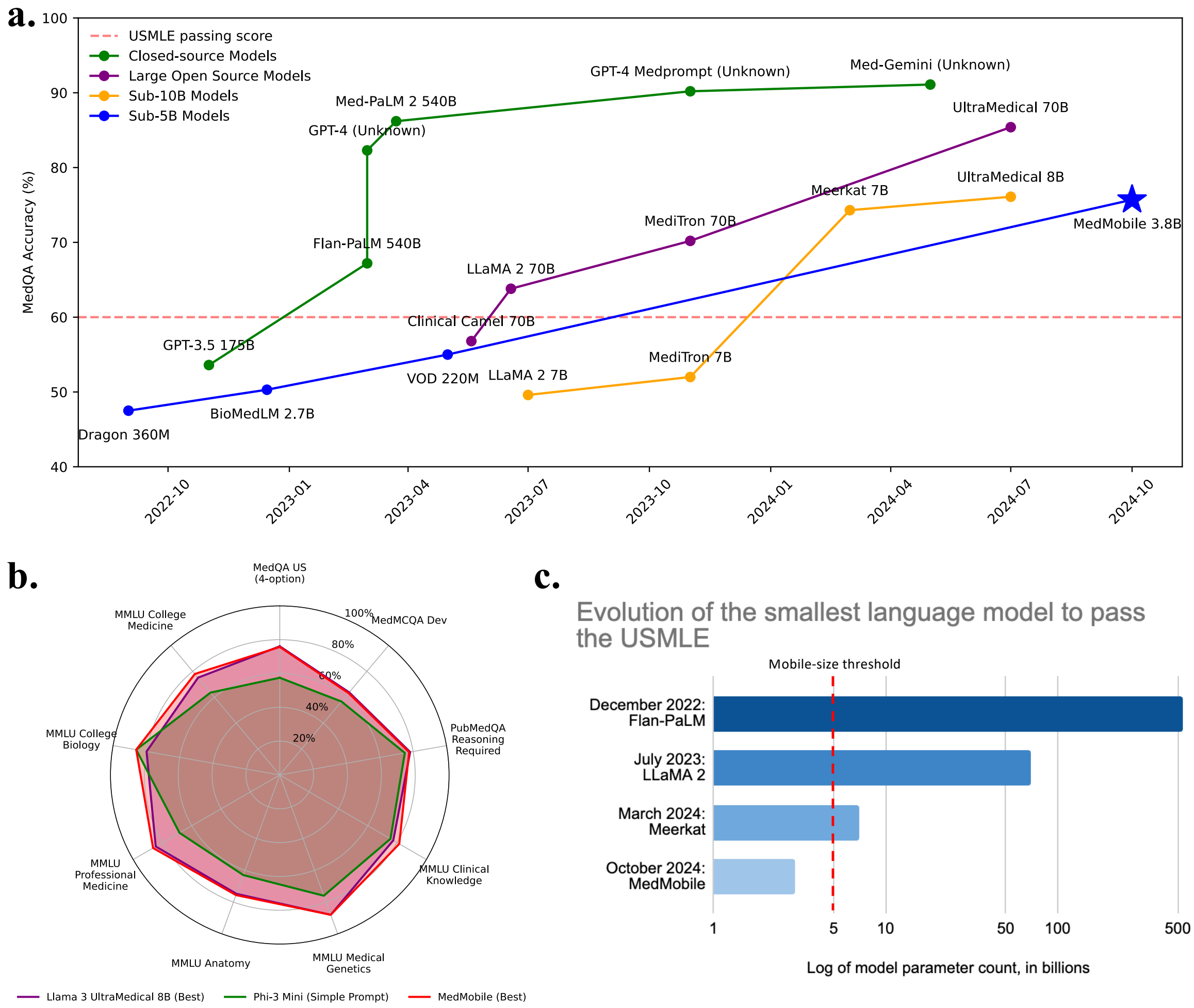

**Figure 4.** Evaluation results of MedMobile. **a.** Depiction of language models in medicine and their accuracy on the MedQA (USMLE). We broadly classify models into four groups, closed-source models, large open-source models, sub-10B parameter language models, and sub-5B parameter language models. **b.** Comparison of the performance of MedMobile to UltraMedical and phi-3-mini across the MultiMedQA. **c.** Illustration of the smallest model of a time that can pass the USMLE (based on the MedQA benchmark) and the log of the model's parameter count.

<u>Data leakage in evaluation methods</u>

The MedQA is a common benchmark and is highly publicized online. As such, it is likely subject to data leakage, especially on models that do not utilize heavy filtering during their training data curation. To address this, we run a second round of testing on the less popular medical benchmark, Medbullets (4-op). All tested models drop dramatically in accuracy when tested on the less popular medical benchmark, Medbullets (4-op), despite still testing on USMLE-style questions (average loss = 14.4±0.25%).

Computational cost estimation

To characterize the computational savings from utilizing smaller models, we conduct a rudimentary analysis of the required costs to deploy and train clinical models (Table 2). MedMobile requires roughly half the memory footprint of contemporary 8 B-parameter medical Llama variants and more than two orders of magnitude less than 400 B-class foundations. This translates into a single-GPU deployment ($0.53/hr), delivering a 48 % reduction in hourly inference cost and a 53 % reduction in GPU VRAM relative to the 8 B baselines (e.g., Meerkat and UltraMedical). Training shows an even starker contrast: MedMobile completes on a relatively modest hardware setup ($1.58/hr), slashing hourly training spend by ~40 % compared to the 8B baselines. Against ultra-large models such as DeepSeek-R1 or Llama-3.1-405B, MedMobile's deployment and training costs are ~98 % lower.

**Table 2.** AWS GPU cluster costs for medical LLM inference and training at FP16 precision. Costs represent minimum on-demand pricing in *us-east-1* region, with inference requiring 20% memory overhead and training using Adam optimizer (4× model size plus 20% overhead).

| | Weights at FP16 (GB) | $VRAM_{infer}$ (GB) | $VRAM_{train}$ (GB) | deploy $/hr | train $/hr |
|---|---|---|---|---|---|
| MedMobile | 7.1 | 8.5 | 34 | $0.53 | $1.58 |
| Llama-3-8B-UltraMedical | 15 | 17.9 | 72 | $1.01 | $2.63 |
| llama-3-meerkat-8b-v1.0 | 15 | 17.9 | 72 | $1.01 | $2.63 |
| DeepSeek-R1-0528 | 641.3 | 769.6 | 3078 | $25.77 | $101.52 |
| Llama-3.1-405B-Instruct | 756 | 907.2 | 3629 | $29.98 | $119.40 |

## Discussion

In this study, we set out to bridge the gap between large, proprietary clinical language models and the pressing need for an accessible, HIPPA-compliant alternative by introducing MedMobile: a compact, open-source model fine-tuned on medical QA and reasoning via knowledge distillation from GPT-4. Our evaluation on the MultiMedQA benchmark reveals that MedMobile achieves expert-level accuracy across diverse clinical tasks and produces transparent, step-by-step chain-of-thought explanations that enhance interpretability in complex patient scenarios. Crucially, despite inherent context-window constraints, MedMobile consistently outperforms a phi-3-mini baseline across varied output-length bins, underscoring the value of domain-specific fine-tuning over conventional prompt-engineering approaches.

MedMobile represents a significant step towards building towards more accessible and secure language models in medicine. Because the model runs comfortably on a single commodity GPU at roughly $0.50 per hour for inference and $1.60 per hour for training, point-of-care or edge deployment becomes economically feasible, cutting operational expenditure by ~40–98 % compared with 8 B- and 400 B-class medical LLMs (Table 2). We demonstrate high levels of proficiency in medical Q&A across a variety of medical tasks in the MultiMedQA. However, we note a decrease in performance based on token output length but not token input length (see Supplemental Figure 1 and Supplemental Figure 2). This phenomenon has been documented in LLMs, where model performance initially increases with longer CoT, but then decreases after a certain CoT length [26]. However, across almost all bins, irrespective of output length, MedMobile outperforms phi-3-mini, highlighting the gain of domain-specific knowledge and the resultant gain in MedMobile's performance on medical tasks.

MedMobile is trained from a dual synthetic and human curated corpus. We opt to not solely rely on synthetic data for fine-tuning medical foundation models. This may limit clinical validity and generalizability, as such data may lack the nuance, variability, and noise present in real-world clinical texts [27,28]. Our hybrid approach grounds the model in some human truth and allows us to scale the training dataset to a greater degree.

Contrary to popular literature, in our final pipeline, we do not utilize many of the prompt engineering approaches that are common for large language models RAG and k-shot prompting. We implemented these techniques (see Supplemental Figure 3), but they did not lead to any significant degree of improvement. We hypothesize that this is mainly driven by the context-window limitations small language models have and note these are interesting barriers to tackle for future research [8].

There are several limitations that apply to our work. While we demonstrate a significant improvement over previous open-source models of MedMobile's size, any-size models still demonstrate superior performance on medical tasks. Thus, ignoring the barriers of subscription fees and issues with uploading classified patient health information, GPT-4o can be used for quick and reliable online inference.

Multiple-choice questions (MCQs) are the prevailing evaluation standard for clinical language models, drawing on their longstanding role in medical licensure and educational assessments to provide objective, reproducible metrics and automated scoring pipelines that facilitate direct model comparisons [13]. This allows for standardization across evaluations between different LLM agents and humans, and an ability to scale evaluation to specific needs without requiring manual qualification of results. Nevertheless, the constrained format of MCQs can limit the expression of nuanced clinical judgment and multistep reasoning, such as the formulation of differential diagnoses or justification of management plans, areas in which open-ended or essay-based assessments offer richer evaluative granularity [29,30].

Widely used datasets such as MedQA are heavily circulated online and likely appear (at least in part) in pre-training corpora of both proprietary and open-source LLMs. This raises the risk that reported gains reflect memorization rather than genuine clinical reasoning [31–33]. To probe this, we repeated our evaluation on the much less publicized Medbullets (4-op) benchmark. The Medbullets benchmark's recency, low crawler exposure, independent leakage testing, and niche visibility make it a viable alternative candidate for testing on USMLE-style questions. All models, including MedMobile, experienced a marked accuracy decline, suggesting that at least part of the performance on MedQA may stem from exposure to test items during pre-training. While our two-benchmark design reduces (but does not eliminate) the leakage concern, we caution that residual overlap, imperfect heuristic filtering, and the absence of a completely sequestered test set still limit confidence in headline accuracy numbers and inferences about out-of-distribution generalization.

Although MedMobile excels on standardized medical exams, as a recent analysis notes, these benchmarks fail to capture the ambiguity, incompleteness, and complexity of real-world clinical data, underscoring the urgent need for rigorous validation in authentic care environments [34–36]. Thus, we note that real-world clinical and patient-facing deployment of MedMobile is yet to be evaluated and is left for future works. Prospective deployment will additionally require compliance with FDA guidance, continuous hallucination monitoring, and HIPAA-conformant logging. Future studies should also perform more comprehensive ablation studies to evaluate various combinations of pipeline components and ordering. This may include systematically identifying and evaluating other compact base models beyond phi-3-mini suitable for edge-inference and medical reasoning. Finally, MedMobile, in its present form, is trained only on language and cannot ingest medical images, an obvious constraint given that many diagnostic workflows are image centric. Yet bedside interpretation of chest radiographs, dermatologic photos, or retinal scans often dictates immediate management. Extending MedMobile beyond text is therefore a direct next step for future, more powerful integration into medical systems.

Future directions

This work can be expanded to vision-language models (VLMs) by building upon Phi-3-vision using this pipeline. VLMs have shown promise for superhuman predictive power and novel pattern recognition but notoriously require extensive training and inference costs due to the larger data sizes associated with high-resolution imaging [37,38]. Using a smaller, domain-specific model, such as MedMobile, serves to combat these rising computational costs. Alongside this rise, we also note the increase in novel imaging methods that provide new dimensional data that machine learning models can leverage, such as photoacoustic imaging providing spectral information to individual voxels or shear wave elasticity imaging providing information about tissue stiffness [39,40]. In light of the increase in imaging data from new dimensions (e.g., spectral data from photoacoustic imaging or tissue stiffness data from shear wave imaging) in these modalities, smaller language models may serve to foster new, cutting-edge insights and patterns that otherwise are hidden from humans, while bolstering quick compute times. In tandem with improvements of imaging modalities, VLM pattern recognition, and the increase in mobile-based ML platforms, such as Apple's new Apple Intelligence [41], we envision a method of use for mobile-sized VLMs centered around accessibility, where doctors and patients can take images with their iPhone and receive insights from an expert-level, fine-tuned LLM, without comprising personal security or requiring extensive computing power.

Recent studies in other domains have also demonstrated effective improvements to benchmark accuracy when using multi-LM agent-based systems [42]. A promising avenue for future research could be utilizing MedMobile as part of a multi-LM system, where problem-solving is divided into multiple iterations of MedMobile. Further distillation of GPT-4 on each agent in such an ensemble may allow for significant improvements to accuracy.

## Conclusion

Expert-level, mobile-sized language models, such as MedMobile, hold promise in low and middle-resource settings due to their reduced compute requirements and quicker inference times [43–45] and also serve to democratize access to the technical capabilities of LLMs beyond the domain of large technology companies and groups with substantial computing budgets. While we develop this work primarily for its impact in the medical domain, mobile-size language models and the related techniques in this work can be applied to any domain to train expert-level mobile assistants. We hope this work and our corresponding open-source code base will contribute to the clinically meaningful development of mobile-sized language models that benefit physicians and patients.

## Data availability

The model weights are available on https://huggingface.co/KrithikV/MedMobile [46]. Icons were sourced from the Noun Project (https://thenounproject.com/).

**Code availability**
The code utilized to generated or analyze our model during the current study are available in the nyuolab/MedMobile GitHub repository, https://github.com/nyuolab/MedMobile.

**Acknowledgements**
EKO is supported by the National Cancer Institute's Early Surgeon Scientist Program (3P30CA016087-41S1) and the W.M. Keck Foundation. We would like to acknowledge Nader Mherabi and Dafna Bar-Sagi, Ph.D., for their continued support of medical AI research at NYU. We thank Michael Constantino, Kevin Yie, and the NYU Langone High-Performance Computing (HPC) Team for supporting computing resources fundamental to our work.

**Author Contributions**
EKO conceptualized and supervised the study. KV designed the MedMobile LLM pipeline. KV, JS, and AA implemented and trained the LLM. KV evaluated and tested the LLM. JS aided with LLM serving and deployment. KV wrote the initial draft of the manuscript. KV, AA, DAA, EKO edited the manuscript. All authors revised and approved the manuscript. KV takes responsibility for the accuracy of the data and the integrity of the analysis (guarantor).

**Disclosures**
EKO has equity in Delvi, MarchAI, and Artisight, income from Merck & Co. and Mirati Therapeutics, employment in Eikon Therapeutics, and consulting for Sofinnova Partners and Google. The other authors have no personal, financial, or institutional interest pertinent to this article.

**Statement of Ethics Approval**
This study does not require ethics approval as no patient or animal data is directly utilized. All datasets used for training are publicly available online and are described in the Methods section.

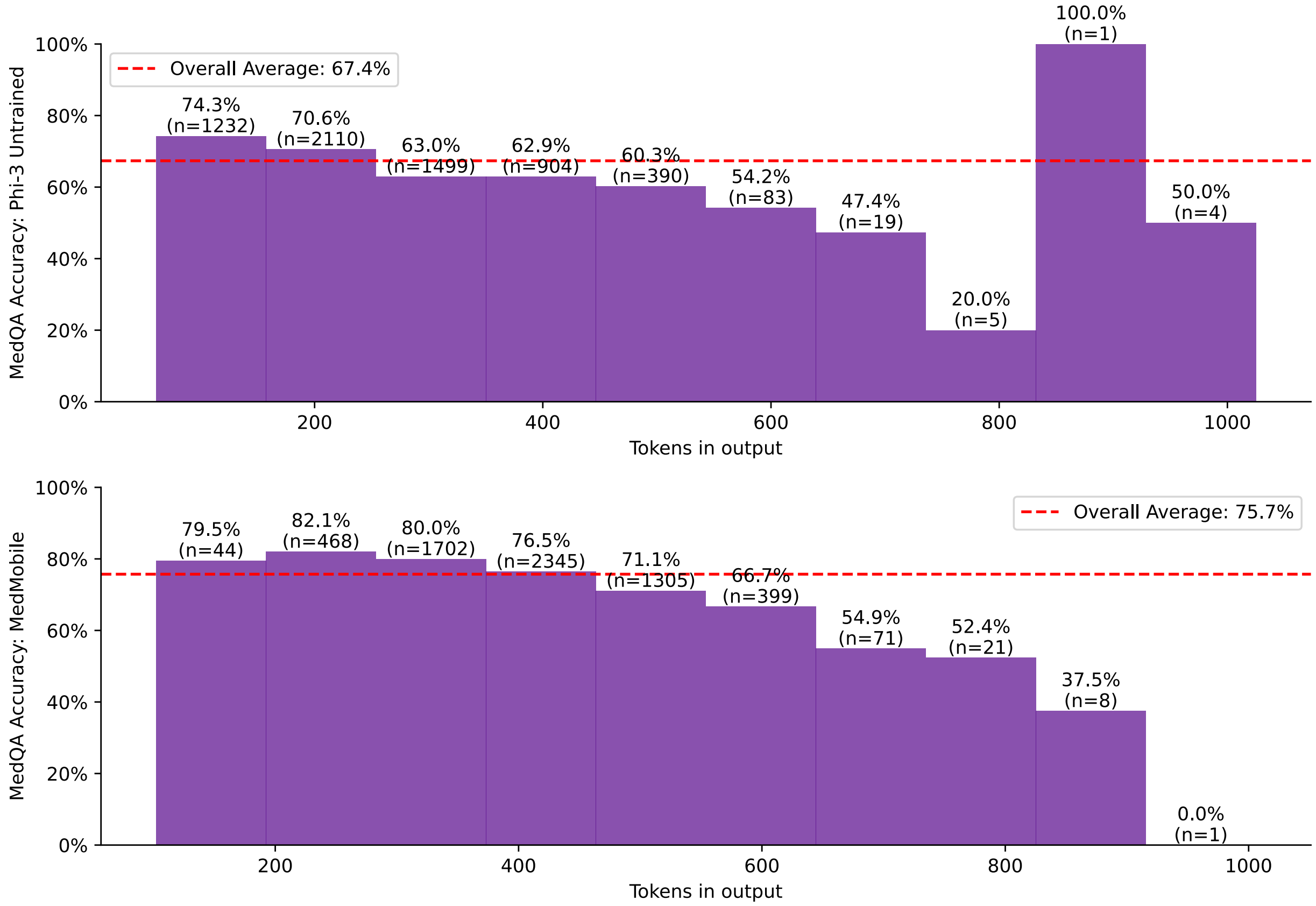

**Supplemental Figure 1.** Comparison of number of output tokens in a response and accuracy on MedQA questions. Each question of the MedQA test set is represented 5x in this figure due to the ensemble performed. Some questions are not included in the plots (< 20) as model response exceeded maximum generation output and an accuracy could not be evaluated. Top panel is a CoT enhanced baseline phi-3-mini model, whereas the bottom panel is our fine-tuned model, MedMobile.

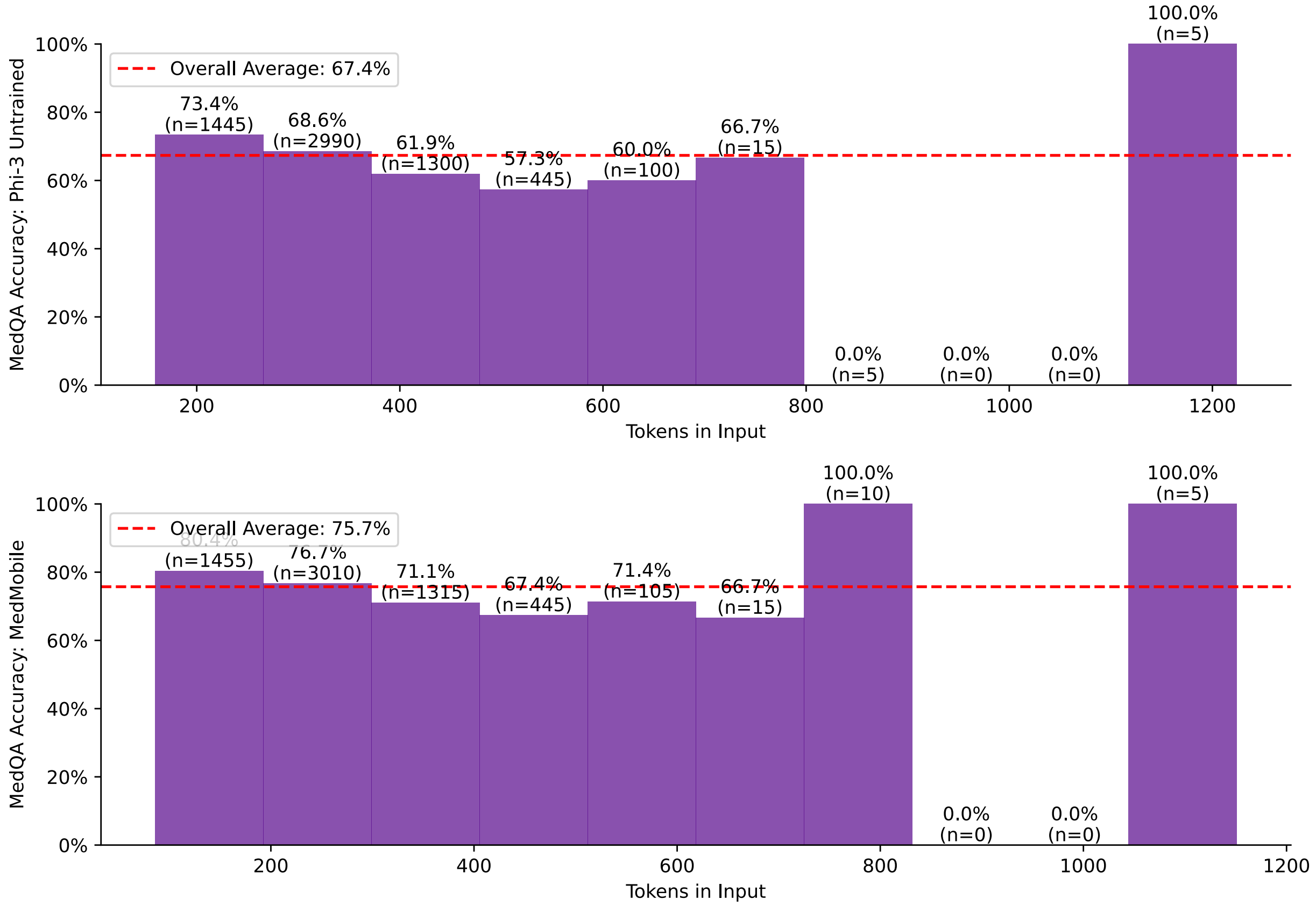

**Supplemental Figure 2.** Comparison of number of input tokens in a response and accuracy on MedQA questions. Each question of the MedQA test set is represented 5x in this figure due to the ensemble performed. Some questions are not included in the plots (< 20) as model response exceeded maximum generation output and an accuracy could not be evaluated. Top panel is a CoT enhanced baseline phi-3-mini model, whereas the bottom panel is our trained model, MedMobile.

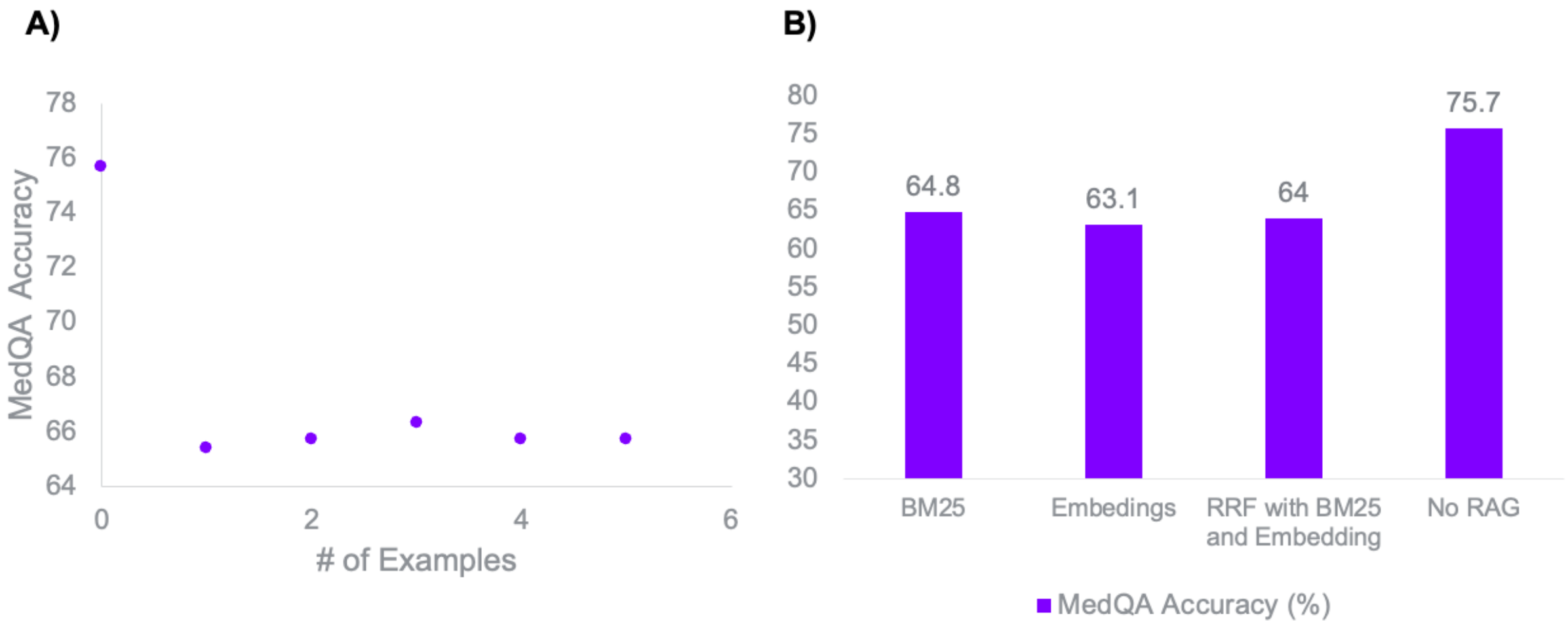

**Supplemental Figure 3.** Panel A) depicts the accuracy of MedMobile on the MedQA relative to the number of k-shot prompting (i.e., number of examples given to the model alongside the evaluation question). Panel B) shows different forms of retrieval for RAG and their resultant effects on the accuracy of MedMobile on the MedQA dataset. To conduct RAG based on vector embeddings, we compute cosine similarity based on MedCPT vectors generation between the question and paragraphs in the textbook. RAG built on BM-25 is developed through the lucine implementation, and selects the paragraph with the highest score for a particular question. While all forms of RAG achieve sub-optimal results, we note that BM25 seemed to affect the model least negatively with the addition of context. The source of information for these evaluations is from Harrison’s Principles of Internal Medicine, 21e [20].